\documentclass[a4paper, amsfonts, amssymb, amsmath, reprint, showkeys, nofootinbib, twoside, superscriptaddress]{revtex4-2}
\usepackage[pdftex, pdftitle={Article}, pdfauthor={Author}]{hyperref}
\usepackage[margin=0.75in]{geometry}

\usepackage{graphicx} 
\graphicspath{ {Figures/} }
\usepackage{epstopdf}
\usepackage{caption}
\usepackage{amsmath}
\usepackage{amsfonts}
\usepackage{amssymb}
\usepackage{bm}
\usepackage{hyperref}
\usepackage{color}
\usepackage{url}
\usepackage{color}
\usepackage{algorithmic}
\usepackage{bbm}
\usepackage{booktabs}
\usepackage{array}
\usepackage[table]{xcolor}
\usepackage{placeins}

\newcommand{\mc}{\mathcal}

\newcommand{\D}{\mathrm{d}}

\begin{document}
\title{$\Gamma$-VAE: Curvature regularized variational autoencoders for uncovering emergent low dimensional geometric structure in high dimensional data}
\date{\today}

\author{Jason Z. Kim}
    \email[Correspondence email address: ]{jk2557@cornell.edu}
    \affiliation{Department of Physics, Cornell University, Ithaca, NY 14853, USA}
\author{Nicolas Perrin-Gilbert}
    \affiliation{Sorbonne Université, CNRS, Institut des Systèmes Intelligents et de Robotique, ISIR, F-75005 Paris, France}
\author{Erkan Narmanli}
    \affiliation{Inserm U830, Institut Curie Research Center, PSL University, Paris, France}
    \affiliation{Department of Translational Research, Institut Curie Research Center, PSL University, Paris, France}
\author{Paul Klein}
    \affiliation{Inserm U830, Institut Curie Research Center, PSL University, Paris, France}
    \affiliation{Department of Translational Research, Institut Curie Research Center, PSL University, Paris, France}
\author{Christopher R. Myers}
    \affiliation{Department of Physics, Cornell University, Ithaca, NY 14853, USA}
    \affiliation{Center for Advanced Computing, Cornell University, Ithaca, NY 14853, USA}
\author{Itai Cohen}
    \affiliation{Department of Physics, Cornell University, Ithaca, NY 14853, USA}
    \affiliation{Kavli Institute at Cornell for Nanoscale Science, Cornell University, Ithaca, NY 14853}
\author{Joshua J. Waterfall}
    \affiliation{Inserm U830, Institut Curie Research Center, PSL University, Paris, France}
    \affiliation{Department of Translational Research, Institut Curie Research Center, PSL University, Paris, France}
\author{James P. Sethna}
    \email[Correspondence email address: ]{sethnap@cornell.edu}
    \affiliation{Department of Physics, Cornell University, Ithaca, NY 14853, USA}


\begin{abstract}
Natural systems with emergent behaviors often organize along low-dimensional subsets of high-dimensional spaces. For example, despite the tens of thousands of genes in the human genome, the principled study of genomics is fruitful because biological processes rely on coordinated organization that results in lower dimensional phenotypes. To uncover this organization, many nonlinear dimensionality reduction techniques have successfully embedded high-dimensional data into low-dimensional spaces by preserving local similarities between data points. However, the nonlinearities in these methods allow for too much curvature to preserve general trends across multiple non-neighboring data clusters, thereby limiting their interpretability and generalizability to out-of-distribution data. Here, we address both of these limitations by regularizing the curvature of manifolds generated by variational autoencoders, a process we coin ``$\Gamma$-VAE''. We demonstrate its utility using two example data sets: bulk RNA-seq from the The Cancer Genome Atlas (TCGA) and the Genotype Tissue Expression (GTEx); and single cell RNA-seq from a lineage tracing experiment in hematopoietic stem cell differentiation. We find that the resulting regularized manifolds identify mesoscale structure associated with different cancer cell types, and accurately re-embed tissues from completely unseen, out-of distribution cancers as if they were originally trained on them. Finally, we show that preserving long-range relationships to differentiated cells separates undifferentiated cells---which have not yet specialized---according to their eventual fate. Broadly, we anticipate that regularizing the curvature of generative models will enable more consistent, predictive, and generalizable models in any high-dimensional system with emergent low-dimensional behavior.

\end{abstract}
\maketitle

Many natural systems with high dimensional states give rise to emergent behaviors that can be described using low dimensional models: the average behavior of $\sim10^{23}$ gas molecules can be described using the ideal gas law, and the expression of $\sim10^{4}$ genes organize along regulatory networks with a relatively sparse number of phenotypes \cite{transtrum2014model}. Successful models of such systems often explain the data using only a few interpretable dimensions \cite{teoh2020visualizing,quinn2019visualizing}, make accurate predictions about new conditions outside of the training data that have not yet been tested, and elucidate fundamental governing principles of the phenotypes that emerge in the low dimensional structure.

The challenge of finding low-dimensional models of high-dimensional data has spanned over a century \cite{pearson1901liii}. Linear projection methods such as PCA provide perfectly consistent embedding dimensions as linear subspaces that span the entire dataset. However, PCA often requires many dimensions to capture the majority of the data variance, making the dimensions difficult to interpret. More modern techniques such as UMAP and variational autoencoders (VAEs) learn low dimensional nonlinear embeddings by using methods that preserve the similarity between data points \cite{mcinnes2018umap,becht2019dimensionality,van2008visualizing,coifman2006diffusion,belkin2003laplacian,moon2017phate,scholkopf1997kernel,williams2000connection,lopez2018deep,kingma2019introduction}. While many of these state-of-the-art methods excel in preserving the topology and local structure of the data, they significantly distort the longer-range data trends, especially in regions where there are no training samples. These distortions limit the interpretability of the embedding dimensions and make it difficult to form accurate predictions on out-of distribution samples, both of which moderate their utility in elucidating the principles underlying emergent behaviors \cite{kobak2021initialization,arvanitidis2017latent}. 

Excitingly, recent work has begun to use tools from differential geometry to more accurately reflect the geometry of the data in the latent representation by measuring and controlling the \textit{curvature} of the manifold \cite{acosta2023quantifying,davidson2018hyperspherical,bose2020latent}. Of particular relevance for the problem of reducing manifold non-linearity is the development of techniques for regularizing estimates of the Ricci and extrinsic curvatures to improve reconstruction accuracy \cite{lee2023explicit}. Their pioneering advance was to stay faithful to the nonlinear geometry of the data rather than relying on the inductive bias of the neural network architecture to model data manifolds \cite{jacot2018neural}. 

Here we build on this approach in two fundamental ways. First, we formulate the extrinsic curvature as a projection of the second derivative vector rather than estimating the full projection operator---resulting in computing an $N$-dimensional vector rather than estimating an $N\times N$-dimensional matrix---where $N$ is the large dimensionality of the data. This formulation allows us to scale to datasets with thousands of features using an exact calculation of curvature rather than an estimate \textit{via} sampling. Second, instead of the Ricci curvature, we regularize the \textit{parameter-effects curvature} \cite{transtrum2011geometry}, which explicitly accounts for the distortion along every direction on the manifold. The parameter-effects curvature is also scalable, exactly calculable, and yields an important theoretical limit: at zero extrinsic and parameter-effects curvature, the second derivative of the manifold in all directions is 0, such that the embedding is analagous to PCA.

To demonstrate its utility, we apply this method to both bulk and single-cell RNA-seq gene expression data. For bulk RNA-seq, we use a joint dataset of both TCGA and GTEx \cite{weinstein2013cancer,lonsdale2013genotype,wilks2021recount3}, and demonstrate the ability to capture the mesoscale organization of healthy and cancer tissues along interpretable axes, and the accurate and fully out-of-distribution prediction of cancer phenotypes. Further, we demonstrate the utility of our approach in single cell RNA sequencing (scRNA-seq) by accurately predicting cell fates from multipotent cells in a lineage-tracing experiment of hematopoietic stem cell differentiation \cite{weinreb2020lineage}. Taken together, our method provides a model manifold through complex and multi-scale data that preserves interpretable dimensions, makes accurate out-of-distribution predictions, and uncovers biologically relevant geometry.

\section{Curvature Regularized VAEs}
The standard formulation of VAEs models the data as being generated from a probability distribution over the data, $\bm{x}_i \in \mc{R}^N$, and a set of unobserved \textit{latent} variables $\bm{z}_j\in\mc{R}^m$. The idea is to find a probability distribution, $p(\bm{x},\bm{z})$, that best matches the distribution of the measured data after marginalizing over the unobserved latent variables,
\begin{align}
    \label{eq:marginal_likelihood}
    p(\bm{x}) = \int p(\bm{x},\bm{z}) \D\bm{z}.
\end{align}
However, the direct optimization of Eq.~\ref{eq:marginal_likelihood} is challenging because it is usually difficult to calculate. 

VAEs use deep neural networks (DNNs) to define an encoder $q_{\bm{\phi}}(\bm{z}|\bm{x})$ and a decoder $\bm{f}_{\bm{\theta}}(\bm{z})$, and optimize a bound on Eq.~\ref{eq:marginal_likelihood}, the Evidence Lower Bound (ELBO):
\begin{align}
    \label{eq:elbo}
    -\underbrace{\mathbb{E}_{q_{\bm{\phi}}(\bm{z}|\bm{x})} ||\bm{x}-\bm{f}_{\bm{\theta}}(\bm{z})||_2^2}_{\mathrm{reconstruction~error}} - \beta D_{KL}(q_{\bm{\phi}}(\bm{z}|\bm{x})||\underbrace{p(\bm{z})}_{\mathrm{prior}}),
\end{align}
where $\bm{\phi}$ and $\bm{\theta}$ are the DNN weights, and the specific form of the reconstruction error is for a Gaussian decoder. The first term is the reconstruction error between the true data point, $\bm{x}_i$, and decoder's reconstruction of that point, $\bm{f}_{\bm{\theta}}(\bm{z}_i)$, from a sample in latent space $\bm{z}_i \sim q_{\bm{\phi}}(\bm{z}|\bm{x}_i)$. The second term is the Kullback-Leibler divergence between the encoded data in latent space, $q_{\bm{\phi}}(\bm{z}|\bm{x}_i)$, and the \textit{prior}, $p(\bm{z})$, which is usually taken to be a standard normal distribution. 

By maximizing Eq.~\ref{eq:elbo} (or minimizing the negative of the ELBO as the loss, $\bar{\mc{L}}$), the encoder makes the latent-space embeddings similar to a standard normal distribution, and the decoder learns a continuous manifold from the latent space to data space that best reconstructs the data. Due to the nonlinearity of the decoder, VAEs often yield sharp variations of this manifold \cite{higgins2016beta}, making the latent representation difficult to interpret. The hyperparmeter $\beta$ serves to regularize the interpretability of the latent space (termed $\beta$-VAE \cite{fil2021beta}), yet does not fully cure the distortion, leading to the dependence of representations on inductive biases and random seeds for initialization \cite{locatello2020sober}. 

To ameliorate this distortion, we desire two properties in our model manifold. First, we want a unit distance anywhere in latent space to correspond roughly to the same distance on the manifold. Hence, we regularize the \textit{parameter-effects curvature} at each sampled point on the manifold, given by
\begin{align}
    \label{eq:parameter_effects}
    \mc{L}_{PE,i} = \sum_{a} g^{\mu\bar{\mu}} g^{\nu\bar{\nu}} \left(\Gamma_{\mu\nu}^{\kappa} \frac{\partial \bm{f}^a_{\bm{\theta}}}{\partial x^{\kappa}}\right) \left(\Gamma_{\bar{\mu}\bar{\nu}}^{\bar{\kappa}} \frac{\partial \bm{f}^a_{\bm{\theta}}}{\partial x^{\bar{\kappa}}}\right),
\end{align}
where $g^{\mu\nu}$ is the inverse of the metric tensor, $\Gamma_{\mu\nu}^{\kappa}$ are the Christoffel symbols, and Eq.~\ref{eq:parameter_effects} is written in Einstein summation notation (repeated Greek indices are summed). At a point on the model manifold, $\bm{f}_{\bm{\theta}}(\bm{z}_i)$, Eq.~\ref{eq:parameter_effects} is the norm of the component of the second derivative of $\bm{f}_{\bm{\theta}}$ that lies tangent to the manifold at that point. Intuitively, Eq.~\ref{eq:parameter_effects} measures by how much a regular grid of points in latent space is distorted by the decoder in data space \cite{transtrum2011geometry}. 

Second, we want straight lines in latent space to correspond roughly to straight lines in data space. Drawing inspiration from prior work, \cite{lee2023explicit,transtrum2011geometry}, we also regularize the \textit{extrinsic curvature} at each point on the manifold, given by
\begin{align}
    \label{eq:extrinsic}
    \begin{split}
    \mc{L}_{EX,i} = \sum_{a} g^{\mu\bar{\mu}} g^{\nu\bar{\nu}} \left(\frac{\partial^2\bm{f}^a_{\bm{\theta}}}{\partial x^\mu \partial x^\nu} - \Gamma_{\mu\nu}^{\kappa} \frac{\partial \bm{f}^a_{\bm{\theta}}}{\partial x^{\kappa}}\right)\\
    \left(\frac{\partial^2\bm{f}^a_{\bm{\theta}}}{\partial x^{\bar{\mu}}\partial x^{\bar{\nu}}} - \Gamma_{\bar{\mu}\bar{\nu}}^{\bar{\kappa}} \frac{\partial \bm{f}^a_{\bm{\theta}}}{\partial x^{\bar{\kappa}}}\right),
    \end{split}
\end{align}
which is the norm of the component of the second derivative of $\bm{f}_{\bm{\theta}}$ that lies perpendicular to the manifold at that point. Intuitively, Eq.~\ref{eq:extrinsic} measures how much the manifold bends in data space. Importantly, we can directly compute and backpropagate on Eq.~\ref{eq:parameter_effects} and Eq.~\ref{eq:extrinsic} instead of estimating them \textit{via} sampling, even if the data live in thousands of dimensions. The final loss that we minimize then is given by
\begin{align}
    \mc{L} = \bar{\mc{L}} + \frac{1}{M}\sum_{i=1}^M (\gamma\mc{L}_{PE,i} + \delta\mc{L}_{EX,i}),
\end{align}
where $i$ indexes points that are uniformly sampled in the latent space based on the covariance of the embedded points. We call this method $\Gamma$-VAE due to the role of the Christoffel symbols in computing curvature.

\begin{figure*}[!ht]
\begin{center}
\includegraphics[width=\textwidth]{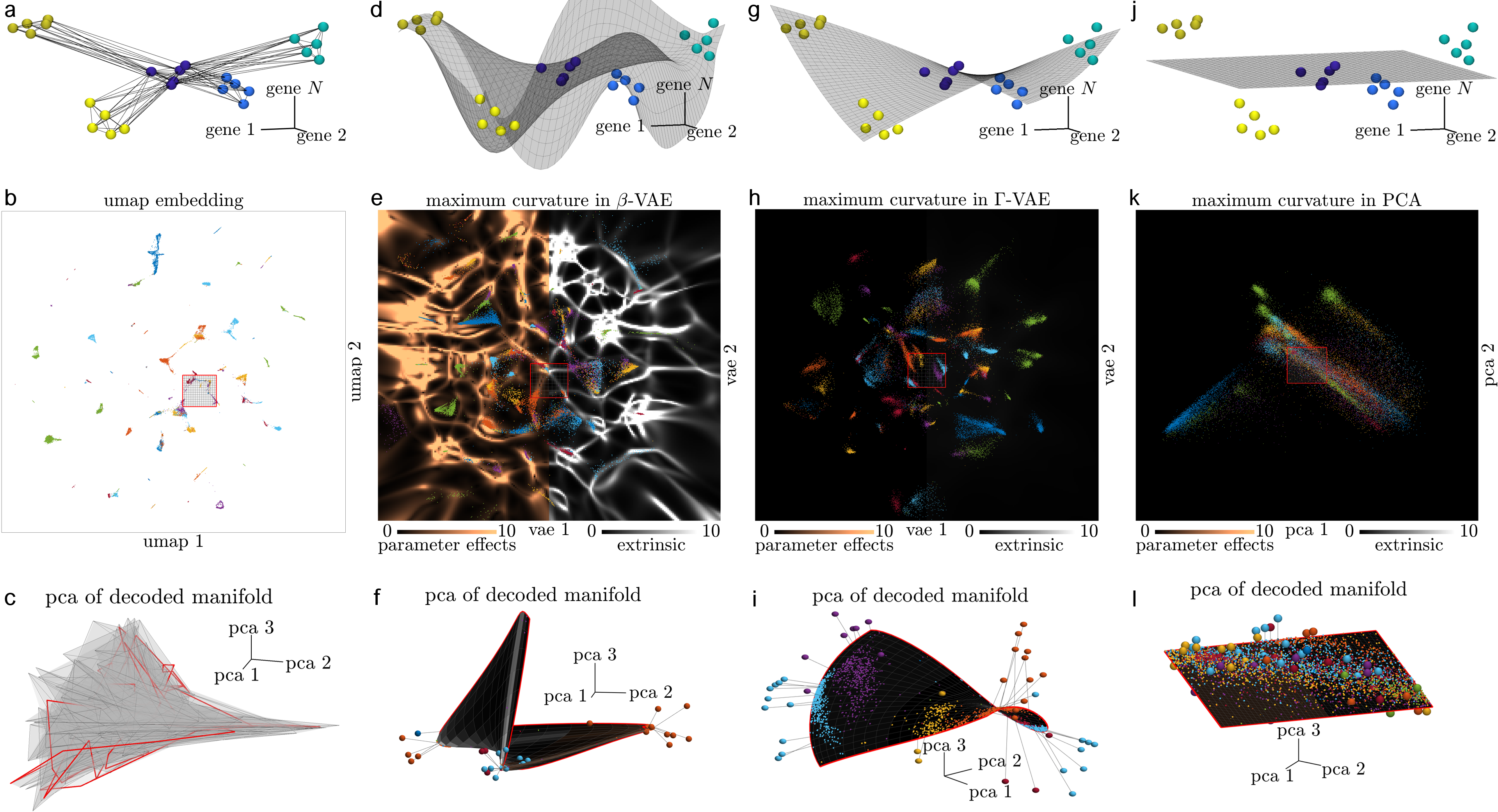}
\caption{\textbf{Explicit regularization of the model manifold curvature}. (\textbf{a}) Schematic of tissues (colored points) connected to nearest neighbors (lines) in the high-dimensional space of gene expression. (\textbf{b}) 2-D UMAP embedding of the joint TCGA + GTEX datasets, with a sampled grid in the center. (\textbf{c}) PCA of the sampled grid projected back into gene space. (\textbf{d}) Schematic of a continuous and differentiable manifold through the tissue samples. (\textbf{e}) Embedding of the dataset using $\beta$-VAE, where the embedding is colored by the maximum parameter-effects curvature (copper, left half), and the maximum extrinsic curvature (grayscale, right), with visibly extreme distortions. (\textbf{f}) PCA of a decoded grid shows a sharply deformed manifold. (\textbf{g}) Schematic of a manifold through tissue samples with less curvature. (\textbf{h}) Embedding of the dataset using $\Gamma$-VAE, demonstrating significantly less parameter-effects and extrinsic curvature, which can be further seen in (\textbf{i}) the PCA of a decoded portion of the latent space. (\textbf{j}). Schematic of a PCA through the datapoints, with (\textbf{k}) a corresponding PCA of the linear embedding and (\textbf{l}) a subset of points.}
\label{fig:figure1}
\end{center}
\end{figure*}

\section{Geometry of model manifolds}

To provide some intuition about the information contained in different low-dimensional embeddings, we consider the manifolds learned on our joint TCGA+GTEx data by four different methods: UMAP, $\beta$-VAE, $\Gamma$-VAE, and PCA. For $n$ data points in $N$-dimensional space, the UMAP algorithm (and related nearest-neighbor embedding methods) constructs a weighted dissimilarity graph between each point and its $k$-nearest neighbors (Fig.~\ref{fig:figure1}a). Then, it uses a force-directed layout to embed the data in a $d$-dimensional space (Fig.~\ref{fig:figure1}b). We observe that UMAP achieves phenomenal clustering between different tissue types. To test whether the embedding contains interpretable geometric information in gene-space, we sample a grid of points in the embedding, decode the sample back into gene-space, plot the PCA of the decoded sample (fig.~\ref{fig:figure1}c), and find that the manifold is quite jagged and heterogeneous.

Next, we train a $\beta$-VAE on the same data, which learns a continuous manifold through the data in gene-space (Fig.~\ref{fig:figure1}d). We use the Softplus activation function to guarantee that the manifold is smooth and differentiable. The $\beta$-VAE embedding also clearly separates the different tissue types (Fig.~\ref{fig:figure1}e), but when we decode a sampled grid on the manifold, we find that the manifold geometry is distorted (Fig.~\ref{fig:figure1}f). We measure this distortion using the norms of the largest parameter-effects curvature (Fig.~\ref{fig:figure1}e, copper, left) and extrinsic curvature (Fig.~\ref{fig:figure1}e, grayscale, right) in any direction along the manifold. As is expected, $\beta$-VAE significantly distorts the regular grid of latent-space points on the manifold, and sharply curves the manifold in orthogonal directions between neighboring tissues (Fig.~\ref{fig:figure1}f).

The idea behind $\Gamma$-VAE is to regularize the parameter-effects curvature (which reduces distortion within the manifold) and the extrinsic curvature (which reduces the bending of the manifold in gene space) to better capture long-range covariance in the data (Fig.~\ref{fig:figure1}g). The corresponding embedding of the TCGA+GTEx data shows the points as being distributed much more uniformly throughout the manifold, with much less parameter-effects and extrinsic curvature (Fig.~\ref{fig:figure1}h). A PCA of a decoded grid sample---containing several of the same tissues as in Fig.~\ref{fig:figure1}e---shows a gently curving manifold that moves roughly in the same direction between multiple tissue types (Fig.~\ref{fig:figure1}i). At the extreme limit of very high curvature regularization, we get PCA that is a linear manifold through the data (Fig.~\ref{fig:figure1}j) with no extrinsic or parameter-effects curvature (Fig.~\ref{fig:figure1}k), where the tissues are not well separated (Fig.~\ref{fig:figure1}l). Hence, $\Gamma$-VAE provides both regularly-spaced and smoothly varying manifolds in gene space with good data separation.

\begin{figure*}[!ht]
\begin{center}
\includegraphics[width=\textwidth]{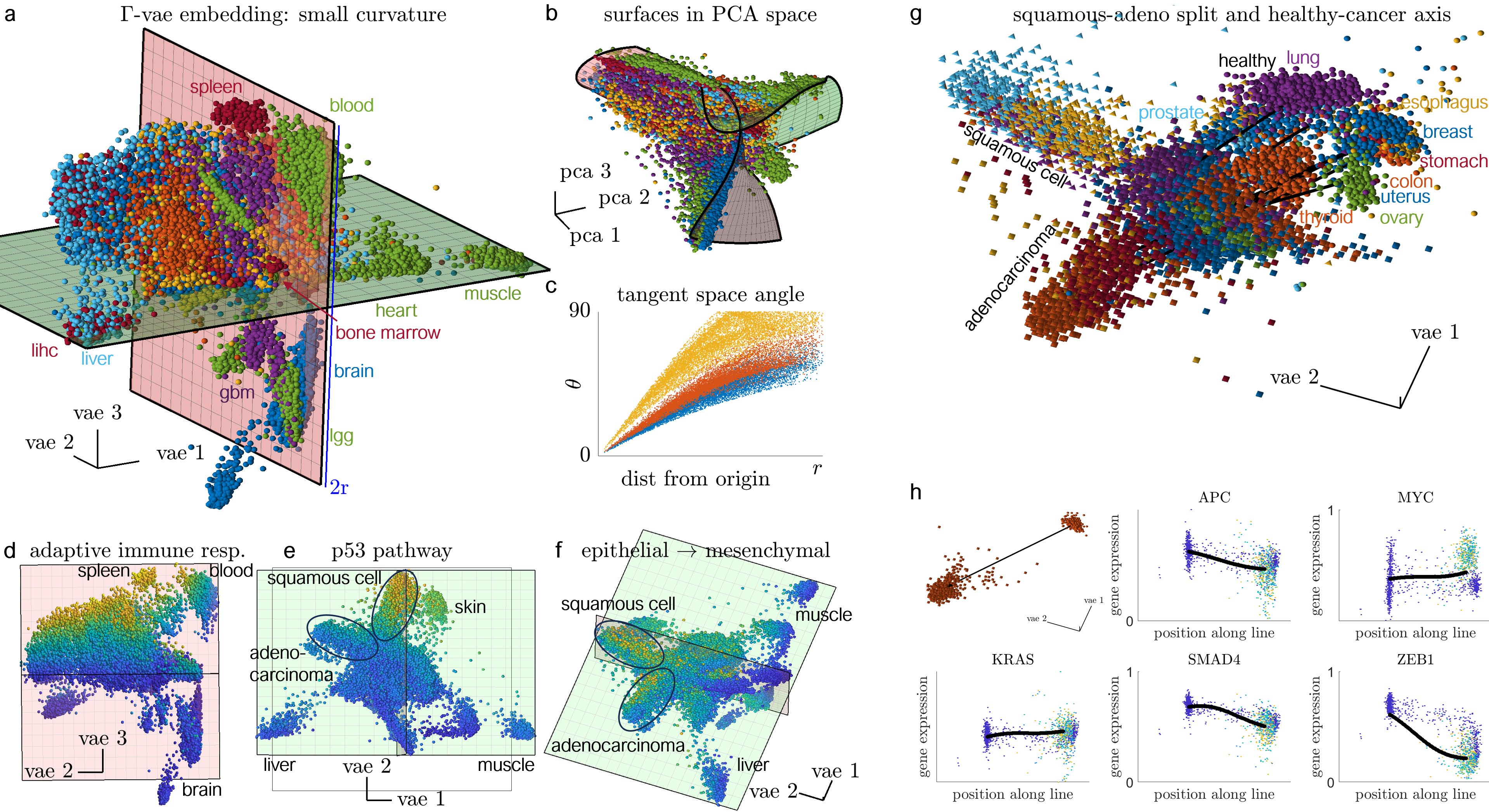}
\end{center}
\caption{\textbf{A geometric 3 dimensional atlas of human tissue and cancer gene expression}. (\textbf{a}) Embedding of the joint TCGA + GTEX dataset into a 3 dimensional latent space of a highly regularized VAE, with a green plane spanning liver and muscle, and a red plane spanning the blood to the brain. (\textbf{b}) Projection of the decoded red and green planes onto a linear PCA of the data, showing the significant yet regularized curvature of the VAE manifold. (\textbf{c}) Angles between the tangent space (in gene space) at the origin \textit{versus} the tangent space radially away from the origin. (\textbf{d}) Plot of the VAE embedding colored by the gene signature of adaptive immune response---which defines the axis from blood to brain---(\textbf{e}) p53 pathway---which defines an axis from bone marrow to the majority of cancer tissues---(\textbf{f}) and epithelial mesenchymal transition---which shows complex spatial gradients in the cancer clusters. (\textbf{g}) Plot of a subset of nine healthy tissues with arrows drawn to their corresponding adenocarcinomas, forming a distinct and uniform cancer axis. The embedding also shows a distinct orthogonal separation of squamous cell carcinomas. (\textbf{h}) Decoded gene trajectories from healthy colon to colon adenocarcinoma for select genes, showing curved, nonlinear pahtways through gene space. }
\label{fig:figure2}
\end{figure*}

\section{A curved 3-dimensional atlas of human gene expression}

With the intuitions from the previous section, we now construct an interpretable, curved 3-dimensional atlas of human gene expression (TCGA+GTEx). We train a $\Gamma$-VAE with a 3-dimensional latent space, and large curvature regularization (Fig.~\ref{fig:figure2}a). We first observe two distinct axes of biological function: one between the liver and muscle (along the green plane), and the second between blood and brain (along the red plane). By ``interpretable,'' we mean that the geometry of this embedding is only mildly distorted in gene space. To demonstrate this distortion, we decode the grid of points in the red and green planes into gene space, and project the decoded planes onto a 3-dimensional PCA of the data (Fig.~\ref{fig:figure2}b), showing a gentle curving of the planes. To further quantify this distortion, we compute the three angles formed by the 3-dimensional tangent spaces of the manifolds between the origin of the embedding and at randomly sampled points (Fig.~\ref{fig:figure2}c). The manifolds do not become orthogonal in any direction until the edges of the latent embedding, demonstrating long-range coherence.

We find that these axes are not only consistent, but also biologically meaningful. We color each tissue in the embedding by the gene signature---the sum of the expression for a subset of genes---associated with adaptive immune response, and find a striking gradient from the blood to the brain (Fig.~\ref{fig:figure2}d). A gene set enrichment analysis (GSEA \cite{subramanian2005gene}) of the tangent vector along this axis from the origin is enriched for processes such as inflammatory response, allograft rejection, and TNFA signaling via NFKB. We find a second global axis from healthy tissues towards the squamous cell carcinomas with a monotonic gradient in the p53 pathway gene signature (Fig.~\ref{fig:figure2}e). The tangent vector along this axis is further enriched in early and late estrogen response, epithelial-mesenchymal transition (EMT), and KRAS signaling. Beyond monotonic gradients, our method also organizes points along spatially heterogeneous patterns, where we find both an EMT axis from healthy tissues to the carcinomas (squamous cell and adenocarcinomas), as well as a gradient for each branch of the carcinomas (Fig.~\ref{fig:figure2}f).

\begin{figure*}[!ht]
\begin{center}
\includegraphics[width=\textwidth]{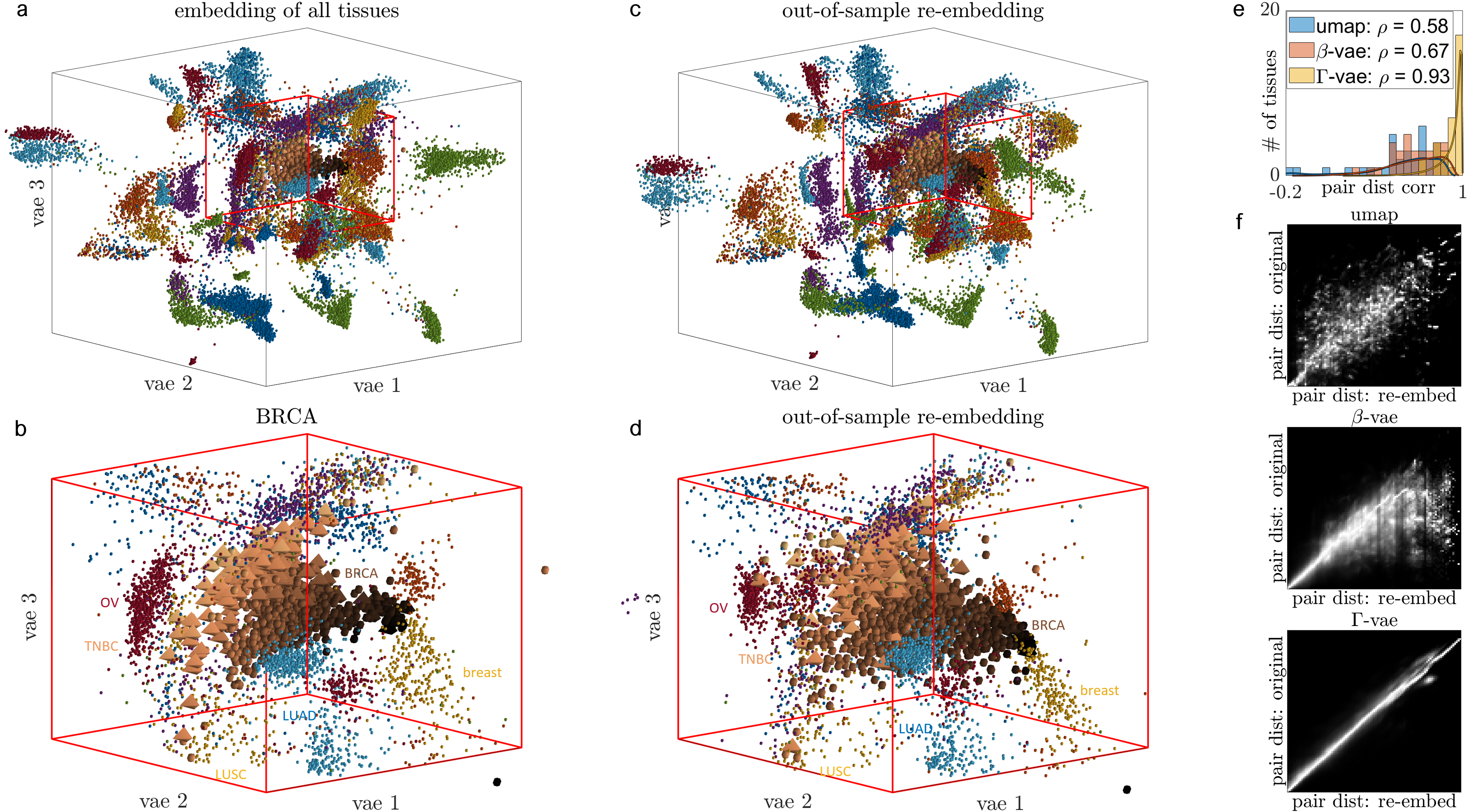}
\end{center}
\caption{\textbf{Out-of-distribution generalizability of reguarlized embeddings to unseen cancers}. (\textbf{a}) A $\Gamma$-VAE embedding of the joint TCGA + GTEX dataset, with a subset of tissues (breast carcinoma, BRCA) colored in a copper gradient based on their distance to healthy breast tissue. (\textbf{b}) a zoomed-in plot of the joint breast and BRCA tissues in the VAE space, where the triple-negative breast carcinomas (TNBC) are plotted as tetrahedra. (\textbf{c}) A regularized embedding trained while leaving out all BRCA and BRCA juxta-tumor samples, with the BRCA tissues re-embedded after training. (\textbf{d}) A zoomed-in plot of the joint breast and re-embedded BRCA tissues that successfully separates TNBC from non-TNBC tissues, where the BRCA tissues are colored the same as in panel (\textbf{b}). (\textbf{e}) Pairwise-distance correlations for UMAP, unregularized VAE, and regularized VAE of each cancer tissue and all other tissues between embeddings where the cancer tissue was included or excluded from the training. (\textbf{f}) Column-normalized density plots of all pairwise distances for each cancer tissue and all other tissues between the original embedding including the tissue, and the out-of-sample embedding that excluded the cancer tissue from training.}
\label{fig:figure3}
\end{figure*}

To better understand the organization of healthy \textit{versus} cancer tissues, we look at a subset of carcinomas and their corresponding healthy tissues (Fig.~\ref{fig:figure2}g). We find that the atlas clearly and geometrically separates the squamous cell carcinomas (tetrahedra) from the adenocarcinomas (cubes) and healthy tissues (spheres). Further, we report a consistent geometric vector from 9 healthy tisues (breast, colon, esophagus, lung, ovary, prostate, stomach, uterus, and thyroid) to their corresponding adenocarcinomas. Our atlas also separates squamous cell carcinomas from adenocarcinomas for cancers originating from the same tissue (e.g. lung squamous cell \textit{versus} lung adenocarcinoma), and that for TCGA categories that contain a mixture of both (e.g. esophageal carcinoma), our atlas correctly embeds the two types as determined by the histology.

Finally, due to VAEs learning an explicit geometric manifold in gene-space, we can study the trajectory of gene expression along axes. In Fig.~\ref{fig:figure2}h, we plot the trajectory of gene expression from healthy colon tissues to colon adenocarcinoma for a subset of genes. Notably, because straight lines in our latent space correspond to gently curved lines in gene space, these gene trajectories have the potential to yield staging information during the progression of tumorogenesis. We note that in this dataset, there is no concrete evidence that these specific trajectories correspond to the process of tumorogenesis. Rather we demonstrate the potential of our method for reconstructing the nonlinear sequence of gene expression \cite{fearon1990genetic}.

\section{Out-of-distribution prediction of cancer phenotypes}

While part of the advantage of constructing geometric manifolds in data space is improved and interpretable visualizations, the manifolds themselves are also a model of the data. We explored the model capabilities by studying the enrichment of tangent vectors (Fig.~\ref{fig:figure2}d-f) and trajectories (Fig.~\ref{fig:figure2}h) of the model manifold. Here, we demonstrate the power of the model to accurately predict out of distribution.

\begin{figure*}[!ht]
\begin{center}
\includegraphics[width=\textwidth]{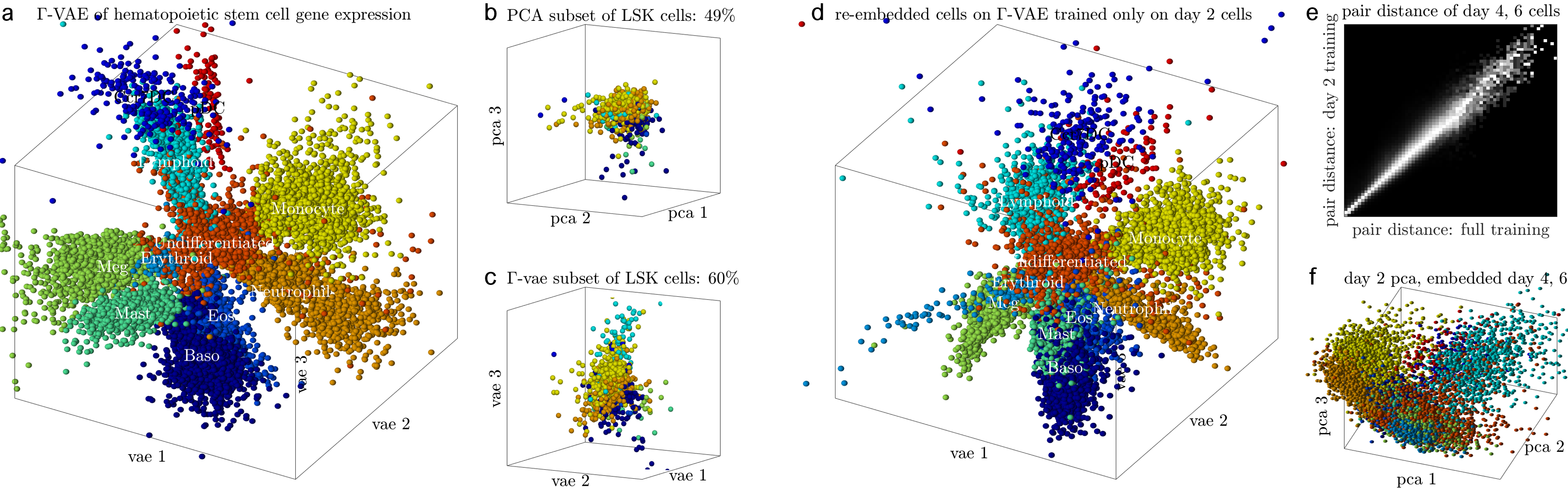}
\end{center}
\caption{\textbf{Cell fate prediction using curved embeddings}. (\textbf{a}) $\Gamma$-VAE embedding of a lineage tracing experiment on hematopoietic stem cell differentiation. (\textbf{b}) Subset of multipotent LSK cells from a PCA of the data, yielding a 49\% classification accuracy of the eventual fate of their daughter cells. (\textbf{c}) The same subset of cells from the cf-VAE of the data, yielding a 60\% classification accuracy of the eventual fate of their daughter cells. (\textbf{d}) re-embedding of all data on the $\Gamma$-VAE manifold trained only on day 2 cells. (\textbf{e}) Column-normalized density plot of pairwise distances between day 4 and 6 cells in the embedding, with the pairwise distances of a $\Gamma$-VAE trained on all data on the x-axis, and the pairwise distances of a $\Gamma$-VAE trained on only day 2 data on the y-axis. (\textbf{f}) Re-embedding of day 4, 6 cells on the PCA projections defined on only the day 2 cells.}
\label{fig:figure4}
\end{figure*}

As a concrete example, we train a $\Gamma$-VAE on all of the TCGA + GTEx data at a lower regularization than in Fig.~\ref{fig:figure2}, which clearly separates most tissues (Fig.~\ref{fig:figure3}a), and color the samples from breast carcinoma (BRCA) in a copper gradient based on their distance from healthy breast tisuses. We choose BRCA because there is significant known within-cancer heterogeneity within BRCA, i.e. triple negative (or basal, TNBC) \textit{versus} HER2+ and hormone receptor positive (or luminal) subtypes \cite{chiu2018integrative}. Zooming into the regions of the embedding near BRCA (Fig.~\ref{fig:figure3}b), we see a clear and expected separation between the non-TNBC BRCA samples (copper spheres) and the TNBC BRCA samples (copper tetrahedra). When we train another $\Gamma$-VAE using a different random number seed after removing all BRCA samples---along with their corresponding juxta-tumor tissues---and re-embed the BRCA samples after training, we find that the geometry of the embedding of all tissues remain relatively fixed, and that the BRCA samples re-embed in the same global location (Fig.~\ref{fig:figure3}c). Zooming into the same region surrounding the re-embedded BRCA samples, we find that the relative distance from the healthy breast tissues (the copper color gradient) is preserved, as is the embedding positions of the non-TNBC and TNBC samples. 

We observed remarkable re-embedding consistency with the out-of-distribution re-embedding of the BRCA samples. To demonstrate the utility of the method more generally, we perform the same procedure of removing each cancer type (along with its corresponding juxta-tumor samples), training a separate $\Gamma$-VAE on the reduced data, and re-embedding the cancer tissues. We quantify re-embedding consistency by first computing the pairwise distance between the out-of-distribution tissues and all other tissues for a $\Gamma$-VAE trained on all of the data, and the pairwise distance for a $\Gamma$-VAE trained on the reduced data (initialized with a different random number seed). We then compute the Spearman rank correlation coefficient between these two sets of pairwise distances for each of the 33 cancer types in TCGA. We find the re-embedding consistency to be significantly higher for $\Gamma$-VAE ($\rho = 0.9$) than for either $\beta$-VAE ($\rho = 0.67$) or UMAP ($\rho = 0.58$) using a non-parmetric Wilcoxon signed rank test ($p<10^{-6}$ for both) (Fig.~\ref{fig:figure3}e). Column-normalized density plots of the pairwise distances across all tissues show the striking ability of $\Gamma$-VAE to preserve local and global embedding distances on out-of-distribution data (Fig.~\ref{fig:figure3}f).

\section{Uncovering and predicting cell fates on curved manifolds}

To demonstrate the utility of $\Gamma$-VAE in another biological context and to extend to single cell RNA sequencing (scRNAseq) data, we look at a lineage tracing experiment for hematopoietic stem cell differentiation \cite{weinreb2020lineage}. In this work, the authors uniquely barcoded a population of hematopoietic stem cells on day 0, collected and sequenced subsets of the population on days 2, 4, and 6, and used the barcodes to trace their differentiation trajectories. Embedding this data using $\Gamma$-VAE, we find that the differentiated cell types are well separated (Fig.~\ref{fig:figure4}a).

The first advantage of embedding on a curved manifold is the ability to separate out the eventual fates of uncommitted cells. A subset of cells collected on day 2 were multipotent LSK cells whose future fates were obtained from their barcode-labelled daughter cells on days 4 and 6. A plot of this LSK subset from a PCA of the data shows poor separation between LSK cells of different eventual fates (Fig.~\ref{fig:figure4}b), with a 49\% classification accuracy using a discriminant classifier on the 3 PCA dimensions. Plotting the same LSK cells in our $\Gamma$-VAE shows significantly better separation between their eventual fates, with a 60\% classification accuracy using a discriminant classifier on the 3 $\Gamma$-VAE dimensions: the same accuracy reported by the original authors using a deep neural network classifier on 447 differentially expressed genes \cite{weinreb2020lineage}.

The second advantage of embedding on a curved manifold is the ability to accurately re-embed out of distribution. We train a second $\Gamma$-VAE on only day 2 cells, and re-embed the day 4 and day 6 cells (Fig.~\ref{fig:figure4}d). Visually, we notice a striking similarity with the embedding trained on all data (Fig.~\ref{fig:figure4}a). We quantify this similarity using the pairwise distances between the day 4 and 6 embedded points, and find a strong correlation in the pairwise distances between the embedding trained on all data, and the embedding trained on only day 2 data (Fig.~\ref{fig:figure4}e) with a Spearman rank correlation coefficient of 0.92. By comparison, the equivalent PCA re-embedding of day 4 and 6 cells on a PCA of only day 2 cells yields significantly worse separation between cell types, and in particular between the differentiated and undifferentiated cells (Fig.~\ref{fig:figure4}f).

\section{Discussion}
The search for interpretable, low-dimensional models has spanned an incredible breadth of time \cite{pearson1901liii,hastie1989principal} subject areas \cite{cunningham2014dimensionality,bruce2002dimensionality}, methodologies \cite{scholkopf1997kernel}, and mathematical objects such as graphs \cite{hamilton2017representation}, mathematical models \cite{transtrum2014model}, and probability distributions \cite{quinn2019visualizing,teoh2020visualizing}. One prevailing paradigm is to search for an appropriate kernel similarity function that serves to untangle the majority of the variance of the data along a few interpretable axes \cite{scholkopf1997kernel} and is the basis for many manifold learning methods today \cite{mcinnes2018umap,becht2019dimensionality,van2008visualizing,coifman2006diffusion,belkin2003laplacian}. In a very real and practical sense, this is also the prevailing mindset in deep learning, where learning in  deep neural networks (DNNs) at a theoretical limit can be thought of as evolution according a kernel function that is determined by the network architecture: the Neural Tangent Kernel (NTK) \cite{jacot2018neural,alemohammad2020recurrent}. The hope is that, at the limit of the ever-increasing quantities of experimental data and conditions \cite{replogle2022mapping,staahl2016visualization} alongside the correct sophisticated DNN architectures \cite{vaswani2017attention}, we can obtain powerful universal representations across a wide variety of data \cite{rosen2023universal}.

In this work, we provide a complementary approach based on a simple idea: that the geometry of the data in natural coordinates is meaningful and worth preserving. This idea is particularly fruitful in heterogeneous atlas data with highly dissimilar clusters where the data topology do not reflect the geometry \cite{lu2023aging,comitani2023diagnostic}, and where statistical modeling methods fail to correctly capture global and meso-scale structure that span between clusters. We find that regularizing the manifold curvature leads to more uniformly distributed embeddings where the genes co-vary across multiple tissue types (Fig.~\ref{fig:figure1}). We also find that regularizing the manifold curvature reveals both global and mesoscale axes of biological function, clear separation of distinct cancer phenotypes, and homogeneous curved gene expression pathways from healthy to cancerous tissues (Fig.~\ref{fig:figure2}). Crucially, our method can accurately predict expression phenotypes not only out of sample, but out of distribution, with far superior consistency than UMAP or $\beta$-VAE (Fig.~\ref{fig:figure3}). Finally, our method can predict future cell fates both by curving the projection of multipotent cells towards their future fate with state-of-the-art classification accuracy in just 3 dimensions, and accurately predict future cell expression from only day 2 data (Fig.~\ref{fig:figure4}). In sum, many biological and emergent processes contain low-dimensional and geometric organization in high-dimensional spaces, and by carefully preserving that geometry, our method constructs more interpretable and predictive low-dimensional models.
~\\

\section{Acknowledgements}
We gratefully acknowledge conversations and feedback from Alexandre Lanau, Phillipe Martin, Jason Cosgrove, Dr. Melody Lim, and Dr. Erich Mueller. JZK was supported by the Bethe/KIC/Wilkins post-doctoral fellowship. This work received financial support from ITMO Cancer of Aviesan within the framework of the 2021-2030 Cancer Control Strategy, on funds administered by Inserm.

\section{References}

\end{document}